\PassOptionsToPackage{table}{xcolor}

\RequirePackage{latexml}
\iflatexml
\documentclass{article}
\usepackage[T1]{fontenc}
\usepackage{graphicx,booktabs,multirow,bm,xcolor,hyperref,cleveref,caption,subcaption}
\usepackage[numbers,sort&compress]{natbib}
\newcommand{\beginappendix}{\appendix}
\else
\documentclass[]{bytedance}
\fi

\usepackage{amsmath,amsfonts,bm}

\def\eqref#1{equation~\ref{#1}}

\def\1{\bm{1}}

\DeclareMathAlphabet{\mathsfit}{\encodingdefault}{\sfdefault}{m}{sl}
\SetMathAlphabet{\mathsfit}{bold}{\encodingdefault}{\sfdefault}{bx}{n}

\usepackage{amsmath}
\usepackage{amssymb}
\usepackage{array}
\usepackage{dcolumn}
\usepackage{algorithm}
\usepackage{algpseudocode}

\iflatexml\else
\usepackage{tikz}
\usetikzlibrary{shapes.geometric,arrows.meta,positioning,fit,backgrounds,calc}
\fi

\usepackage{xspace}
\usepackage{enumitem}
\usepackage{wrapfig}
\usepackage{url}
\iflatexml\else
\usepackage{needspace}
\fi

\newcommand{\TBD}{\,$-$\,}
\newcommand{\myparatight}[1]{\smallskip\noindent{\bf {#1}:}~}

\definecolor{hexec}{HTML}{B0413E}
\definecolor{htool}{HTML}{3B8A4C}
\definecolor{hctx}{HTML}{3B6FB0}
\definecolor{hstate}{HTML}{B9A04F}
\definecolor{hlife}{HTML}{E07020}
\definecolor{heval}{HTML}{6F5BA8}
\definecolor{hmodel}{HTML}{222222}
\definecolor{lightgray}{HTML}{F2F2F2}
\definecolor{medgray}{HTML}{D9D9D9}
\definecolor{seedaccent}{HTML}{2E5AA8}
\definecolor{specbg}{HTML}{FAFAF8}
\iflatexml
\newenvironment{specbox}{\begin{quote}}{\end{quote}}
\else
\newtcolorbox{specbox}{
  enhanced, breakable,
  colback=specbg,
  colframe=medgray,
  boxrule=0.5pt, arc=1.5mm,
  borderline west={2pt}{0pt}{seedaccent},
  left=3mm, right=2.5mm, top=0.5mm, bottom=0.5mm,
  fontupper=\small,
  before skip=8pt, after skip=8pt
}
\fi
\definecolor{accent}{HTML}{1F77B4}
\definecolor{tabhdr}{HTML}{2E2E2E}
\definecolor{tabrow}{HTML}{F4F6F9}

\iflatexml

\else

\fi

\title{\textsc{Axion 1.0}: Open Agent Foundation Models, Simplified}

\iflatexml
\author{Yuqi Jia\textsuperscript{1}, Yunjie Tian\textsuperscript{2} (Project Lead), Jiayu Zheng\textsuperscript{2}, Yijun Liang\textsuperscript{3}, Jingchen Sun\textsuperscript{4}, Mengyu Zhao\textsuperscript{2}, Di Fu\textsuperscript{2}\\
{\small \textsuperscript{1}Duke University; \textsuperscript{2}ByteDance; \textsuperscript{3}University of Maryland, College Park; \textsuperscript{4}University at Buffalo}}
\date{}
\else
\renewcommand{\authorlist}{\begin{tabular}{@{}c@{}}
    \authorformat[1]{Yuqi Jia},\enspace
    \authorformat[2,\dagger]{Yunjie Tian},\enspace
    \authorformat[2]{Jiayu Zheng}\\[4pt]
    \authorformat[3]{Yijun Liang},\enspace
    \authorformat[4]{Jingchen Sun},\enspace
    \authorformat[2]{Mengyu Zhao},\enspace
    \authorformat[2]{Di Fu}
  \end{tabular}}
\affiliation[1]{Duke University}
\affiliation[2]{ByteDance}
\affiliation[3]{University of Maryland, College Park}
\affiliation[4]{University at Buffalo}

\contribution[\dagger]{Project Lead}

\abstract{
We introduce \textbf{Axion 1.0}, a fully open agent foundation model that combines strong and broad agentic capabilities with a remarkably simple training recipe.
Axion 1.0 is trained in only three stages: two lightweight supervised fine-tuning stages followed by reinforcement learning, without dedicated mid-training, direct preference optimization (DPO), or on-policy distillation (OPD).
The first SFT stage develops strong self-contained reasoning through model-aware data construction, where we synthesize challenging problems and adversarially strengthen them based on the base model's own successful solution strategies.
The second stage extends these capabilities to tool-dependent reasoning through a scalable pipeline for constructing diverse and automatically verifiable tasks grounded in real-world data sources and deterministic computation.
Finally, reinforcement learning further strengthens long-horizon reasoning and interaction with external tools and environments.
Despite this compact training recipe, Axion 1.0 achieves strong performance across a broad suite of agentic benchmarks and is competitive with leading commercial models on several of them.
Equally importantly, Axion 1.0 is open beyond model weights: we release the complete recipe, including data construction, generation, filtering, verification, prompts, and training code, enabling the entire development process to be reproduced and extended.
Axion 1.0 demonstrates that strong agent foundation models can be both \emph{open} and \emph{simple}, offering a practical and reproducible path toward advancing open agentic intelligence.
}

\checkdata[Project]{\url{https://axion1-0.github.io/}}
\checkdata[Code]{\url{https://github.com/bytedance/axion1_0}}

\fi

\begin{document}
\maketitle
\iflatexml
\abstract{
We introduce \textbf{Axion 1.0}, a fully open agent foundation model that combines strong and broad agentic capabilities with a remarkably simple training recipe.
Axion 1.0 is trained in only three stages: two lightweight supervised fine-tuning stages followed by reinforcement learning, without dedicated mid-training, direct preference optimization (DPO), or on-policy distillation (OPD).
The first SFT stage develops strong self-contained reasoning through model-aware data construction, where we synthesize challenging problems and adversarially strengthen them based on the base model's own successful solution strategies.
The second stage extends these capabilities to tool-dependent reasoning through a scalable pipeline for constructing diverse and automatically verifiable tasks grounded in real-world data sources and deterministic computation.
Finally, reinforcement learning further strengthens long-horizon reasoning and interaction with external tools and environments.
Despite this compact training recipe, Axion 1.0 achieves strong performance across a broad suite of agentic benchmarks and is competitive with leading commercial models on several of them.
Equally importantly, Axion 1.0 is open beyond model weights: we release the complete recipe, including data construction, generation, filtering, verification, prompts, and training code, enabling the entire development process to be reproduced and extended.
Axion 1.0 demonstrates that strong agent foundation models can be both \emph{open} and \emph{simple}, offering a practical and reproducible path toward advancing open agentic intelligence.
}

\par\noindent Project: \url{https://axion1-0.github.io/}
\par\noindent Code: \url{https://github.com/bytedance/axion1_0}
\par
\fi

\begin{figure}[t]
\centering
\includegraphics[width=\linewidth]{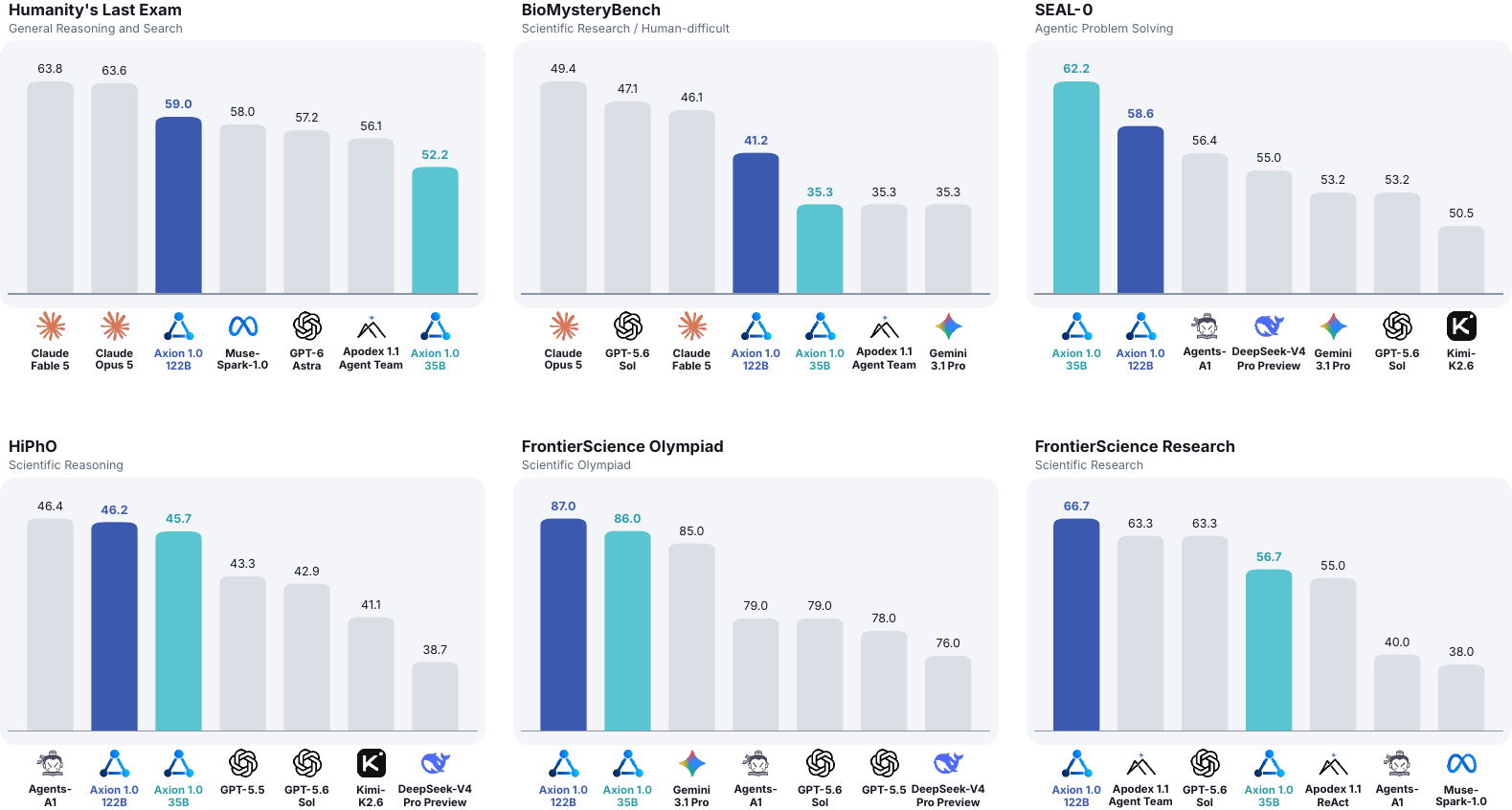}
\caption{\textbf{Performance comparison of Axion 1.0 with leading open and proprietary models} across six challenging benchmarks, spanning general reasoning, scientific reasoning, search, and agentic problem solving. \textcolor[HTML]{1F3D8E}{Blue} and \textcolor[HTML]{74C1D0}{Teal} bars represent Axion 1.0-122B and Axion 1.0-35B, respectively, while gray bars denote competing models. Despite its simple three-stage training recipe, Axion 1.0 achieves competitive or superior performance against frontier models across diverse reasoning and agentic tasks.}
\label{fig:overall_comparison}
\end{figure}

\section{Introduction}
\label{sec:intro}

Large language models are rapidly evolving beyond response generation into agents capable of reasoning, acting, and interacting with external tools and environments~\citep{yao2022react, schick2023toolformer}. This transition is already evident in widely deployed commercial AI systems, such as ChatGPT~\citep{openai2025chatgptagent, openai2026gpt6astra} and Claude~\citep{anthropic2025claude4, anthropic2026claudeopus55}, which increasingly operate as agentic systems rather than relying solely on the outputs of standalone language models. By integrating capabilities such as web search, code execution, and tool invocation, these systems can tackle complex tasks through extended, multi-step trajectories: decomposing objectives, acquiring information, taking actions, interpreting observations, and iteratively refining their solutions. This evolution is also driving the development of \emph{agent foundation models}, where reasoning, tool use, and sustained interaction become intrinsic model capabilities rather than functionality provided solely by external scaffolding. Recent open and proprietary models have demonstrated remarkable progress across reasoning, search, coding, and other tool-augmented tasks~\citep{miromind2026mirothinker17, xie2026quest, agentsa1team2026agentsa1, apodex2026apodex11, moonshot2026kimik26, deepseek2026v4, google2026gemini31pro, openai2026gpt6astra, anthropic2026claudeopus55}.

At the same time, the training pipelines behind these capabilities are becoming increasingly elaborate.
Modern agent models often rely on continued or mid-training, multiple supervised stages, specialized capability-specific data mixtures, preference optimization, reinforcement learning, and additional post-training procedures~\citep{team2025tongyi, miromind2026mirothinker17, xie2026quest, agentsa1team2026agentsa1, apodex2026apodex11}.
While individual components may be well motivated, their accumulation makes it increasingly difficult to understand which ingredients are truly necessary.
Such complexity also raises the cost of reproduction, obscures the sources of capability gains, and makes training recipes harder to transfer across models.
Moreover, many ``open'' models release their weights while leaving substantial parts of the development process---including data construction, filtering, verification, prompts, and training procedures---unavailable.

In this work, we ask a simple question:
\emph{how much of this complexity is actually necessary to build a strong and broadly capable agent foundation model?}
We introduce \textbf{Axion 1.0}, a fully open agent foundation model developed with a deliberately simple training recipe.
Axion 1.0 is trained in only three stages: two lightweight supervised fine-tuning stages followed by reinforcement learning, without dedicated mid-training, direct preference optimization (DPO), or on-policy distillation (OPD).
The first SFT stage develops strong self-contained reasoning, the second extends these capabilities to reasoning that requires interaction with external tools, and reinforcement learning further strengthens long-horizon reasoning and agent--environment interaction.

A central principle behind Axion 1.0 is that simplifying the training pipeline does not require simplifying the learning problem. Instead, we place greater emphasis on the construction of high-quality and challenging training data.
For self-contained reasoning, we develop a model-aware data construction process that synthesizes difficult problems and further strengthens them according to the base model's existing capabilities.
For tool-dependent reasoning, we construct diverse tasks grounded in real-world data sources and deterministic computation, enabling scalable training with automatically verifiable outcomes.
Together, these two stages establish a progression from independent reasoning to grounded interaction, while reinforcement learning further consolidates these capabilities over longer trajectories.
We describe the corresponding data construction and training procedures in detail in later sections.

Despite its compact training recipe, Axion 1.0 develops broad and strong agentic capabilities.
As summarized in Figure~\ref{fig:overall_comparison}, we evaluate Axion 1.0 across a diverse suite of benchmarks covering reasoning, tool use, information seeking, coding, and long-horizon agentic interaction.
Axion 1.0 achieves competitive performance against leading open models and several frontier commercial models across these evaluations.
These results suggest that strong agentic capability need not depend on an increasingly fragmented sequence of specialized training stages.

Axion 1.0 is also designed to be open beyond model weights.
We release the complete development recipe, including data construction, generation, filtering, verification, prompts, and training code.
Our goal is not only to make the final model accessible, but to make its development process reproducible, inspectable, and extensible.
This distinction is particularly important for agent foundation models, whose capabilities often depend as much on training data and interaction design as on the model architecture itself.

Our main contributions are summarized as follows:
\begin{itemize}
    \item We introduce \textbf{Axion 1.0}, a fully open agent foundation model trained with a simple three-stage pipeline consisting of two lightweight SFT stages followed by reinforcement learning, without dedicated mid-training, DPO, or OPD stages.

    \item We develop effective data construction pipelines for both \textbf{self-contained reasoning} and \textbf{tool-dependent reasoning}, combining model-aware task strengthening with scalable construction of diverse and automatically verifiable agentic tasks.

    \item We show that this compact recipe produces \textbf{strong and broad agentic capabilities}, with Axion 1.0 achieving competitive performance against leading open models and several frontier commercial models across a broad benchmark suite.

    \item We release the \textbf{complete development recipe}---including data construction, generation, filtering, verification, prompts, and training code---to support reproducible and extensible research on agent foundation models.
\end{itemize}

Overall, Axion 1.0 demonstrates that strong agent foundation models can be both \emph{simple to build} and \emph{open to reproduce}.
Rather than continually expanding the training stack, our results suggest that carefully constructed data, a clear progression from reasoning to interaction, and reinforcement learning over agentic trajectories can provide a practical path toward capable and reproducible open agents.

\section{Data Construction}
\label{sec:axion1_0}
\label{sec:data_construction}

\begin{figure*}[t]
\centering
\includegraphics[width=\textwidth]{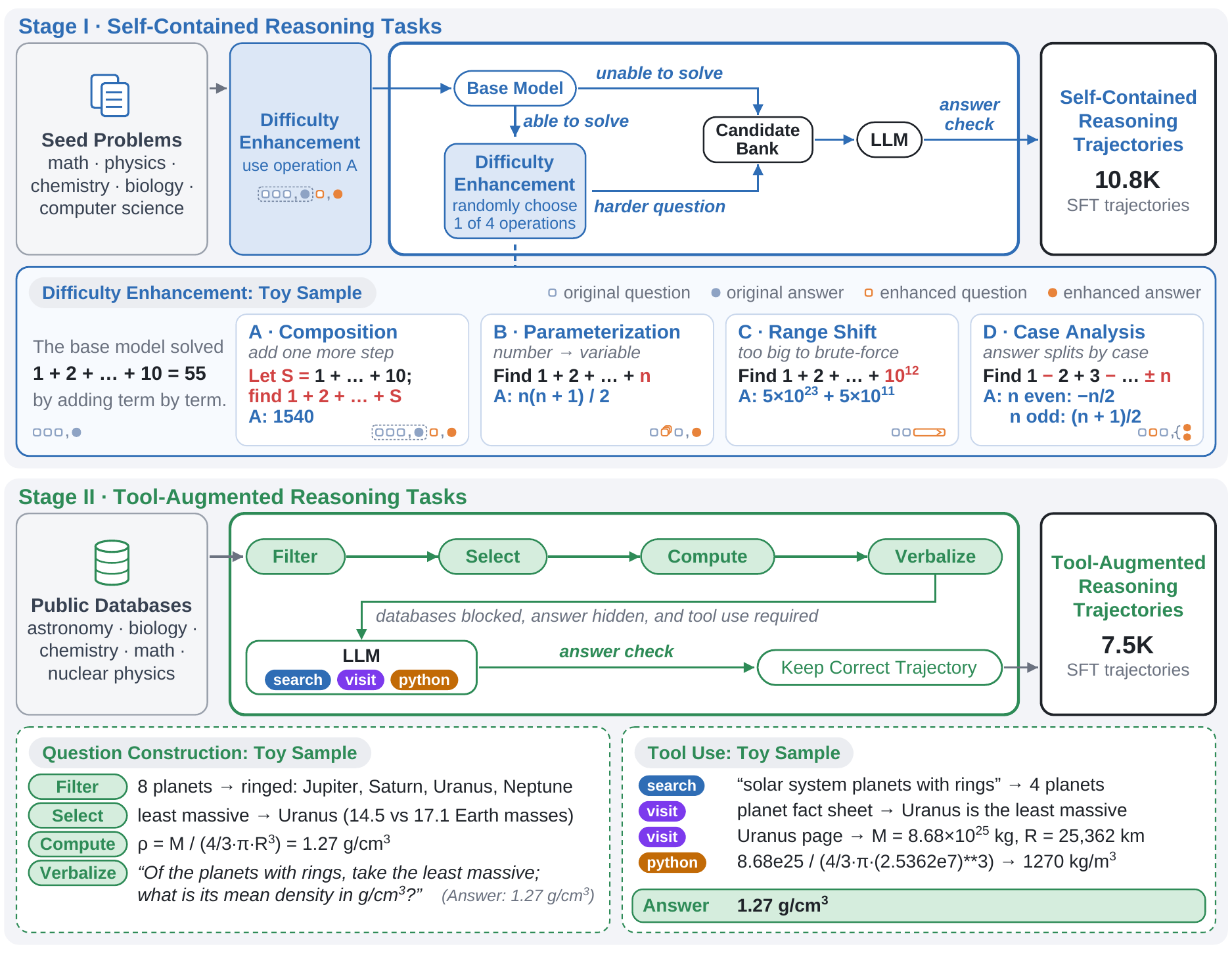}
\caption{\textbf{Overview of the two-stage data construction pipeline.}
\textbf{Stage I} constructs self-contained scientific reasoning problems from public seeds. An initial composition step increases their difficulty; problems solved by the base model are further strengthened using one of four transformations (A--D). An LLM then generates solutions, from which correct reasoning trajectories are retained.
\textbf{Stage II} constructs tool-dependent questions through deterministic filtering, selection, and computation over structured data sources. The construction-time answers and direct database access are withheld during solving, and only verified tool-use trajectories are retained for SFT.}
\label{fig:data_construction}
\end{figure*}

We construct two complementary SFT datasets: one for \emph{self-contained reasoning} and the other for \emph{tool-dependent reasoning}. The first teaches the model to solve challenging problems using the information provided in the question, whereas the second teaches it to acquire missing information through external tools and reason over the retrieved evidence. Both pipelines separate \emph{problem construction} from \emph{trajectory collection}: we first establish a question and a verifiable reference answer, then retain only correctly solved trajectories for training. Figure~\ref{fig:data_construction} summarizes the process.

\subsection{Self-Contained Reasoning Tasks}
\label{sec:stage1}

\myparatight{Seed problems and initial difficulty enhancement}
We collect question--answer pairs from public course materials, competitions, and reasoning datasets spanning mathematics, physics, chemistry, biology, and computer science. Rather than training directly on these potentially familiar problems, we use them as seeds for generating new, more challenging questions.
For each seed, a generator LLM first applies \emph{composition} (operation A): it identifies a quantity established by the seed's solution and constructs a new, self-contained question that requires this quantity as an essential intermediate result. The new question must have an unambiguous, checkable final answer and must not reveal the original seed or its answer. This produces the initial pool of enhanced problems.

\myparatight{Base-model feedback and targeted enhancement}
We evaluate each validated enhanced problem with the base model, allowing Python execution but disabling search and webpage visits. A problem the base model answers incorrectly is added directly to the candidate question bank. If the base model answers correctly, we provide its successful solution to the generator LLM and randomly apply \emph{one} of four difficulty-enhancement operations:
\begin{itemize}
    \item \textbf{A. Composition:} Add a further dependent reasoning step whose solution requires the answer to the current problem as an intermediate result, forming a new question with a new final answer.
    \item \textbf{B. Parameterization:} Replace a fixed value with a symbolic parameter, requiring a general expression rather than a single numerical result.
    \item \textbf{C. Range Shift:} Enlarge or shift the range of a parameter so that simple enumeration or a shortcut used in the original solution is no longer sufficient.
    \item \textbf{D. Case Analysis:} Introduce conditions under which the correct answer differs across cases, requiring the model to identify and resolve each case.
\end{itemize}
The transformed question is added to the candidate bank after answer verification; we do not require the base model to fail a second test before admitting it. In particular, composition can appear twice: once to create the initial problem from the seed, and again as a possible enhancement when the base model solves that problem. The objective is to increase the reasoning required, not merely the length or numerical precision of a question.

\myparatight{Reference-answer verification}
Every newly generated or rewritten problem is checked independently of the base model's prediction. GPT-5.5~\citep{openai2026gpt55} independently solves the problem three times. We accept the reference answer only when all three solutions agree and their derivations pass correctness checks; ill-posed or underspecified problems are discarded. Importantly, an answer is re-established after each rewrite rather than inherited from the previous question. The resulting bank therefore combines initially enhanced problems the base model could not solve with further-strengthened versions of problems it could solve.

\myparatight{Trajectory collection and filtering}
A separate LLM solves the candidate questions with Python available and external retrieval disabled. Each trajectory contains the model's reasoning, any Python calls and their outputs, and the final answer. We compare the final answer against the verified reference and retain only correct, complete trajectories with valid tool interactions and output format; duplicate questions are removed. The resulting self-contained reasoning dataset is
\begin{equation}
\mathcal{D}_{\mathrm{sc}}=\bigl\{(x_i,\tau_i,\varnothing,y_i)\bigr\}_{i=1}^{M_{\mathrm{sc}}},
\label{eq:axion_self_contained_data}
\end{equation}
where $x_i$ is the question, $y_i$ is its reference answer, $\tau_i$ is the retained solution trajectory, and $\varnothing$ indicates that no separate programmatic intermediate verification targets are attached.

\subsection{Tool-Dependent Reasoning Tasks}
\label{sec:stage2}

Unlike Stage I, Stage II primarily targets problems whose answers depend on facts that are not provided in the question. We assemble structured, publicly available data sources covering domains such as biology, chemistry, mathematics, geography, and aviation. Each source supplies records with identifiable entities, attributes, and associated values. We use deterministic programs to construct questions and compute their reference answers, ensuring that the tasks are verifiable before an LLM attempts them. A smaller set of mathematical and algorithmic tasks is generated through deterministic computation and checked by independent implementations.

\myparatight{Filter--select--compute construction}
For database-grounded tasks, each question is built through four steps:
\begin{enumerate}
    \item \textbf{Filter:} Apply explicit conditions to the source records to produce a candidate set of entities.
    \item \textbf{Select:} Identify exactly one entity by a criterion such as the largest or smallest value of an attribute within that set.
    \item \textbf{Compute:} Retrieve additional attributes of the selected entity and calculate a derived quantity using a deterministic formula, including any necessary unit conversions.
    \item \textbf{Verbalize:} Use an LLM to turn the selection conditions and requested computation into one natural-language question, while preserving the programmatically determined reference answer.
\end{enumerate}
For example, we can filter for planets with rings, select the planet with the smallest mass, and ask for its mean density. The solver must first identify Uranus, then find its mass and radius, and finally compute $\rho=M/(4\pi R^3/3)$, yielding approximately $1.27\,\mathrm{g/cm^3}$. This illustrates how a single question combines entity identification, factual retrieval, and computation rather than asking for a directly memorized fact.

\myparatight{Deterministic validation and question formulation}
At construction time, the program checks that each filter produces valid records, the selection criterion identifies a unique entity, and the required attributes are available. When selecting an extreme value, we also require a sufficient margin over the next candidate to avoid ambiguous or unstable answers. For tasks spanning multiple databases, entity identifiers and cross-references connect successive retrieval steps; the output of one step determines the next. Deterministic calculations establish the final answer, with independent implementations used for computational tasks when applicable.
The verbalization LLM may change phrasing but must preserve all selection constraints, values, units, and requested precision. We check the rewritten question for dropped constraints and accidental disclosure of the selected entity or answer. When possible, we also avoid naming the source database, so the question does not reveal a direct lookup path.

\myparatight{Enforcing tool dependence}
The databases used to construct the questions and their stored answers are available to the construction pipeline but are not exposed to the model during trajectory collection. In particular, we block direct access to the construction-time database tables while allowing general-purpose search, webpage visits, and Python. The LLM must therefore recover the necessary facts from accessible external information and combine them to reach the answer. We additionally probe database-grounded questions without retrieval tools and remove those that can already be answered correctly in a closed-book setting. These measures focus the supervision on information seeking and evidence-based computation rather than recall of the reference answer.

\myparatight{Trajectory collection and filtering}
An LLM solves each retained question using search, webpage visits, and Python as needed. We compare its final answer with the reference answer computed during construction, using exact matching or task-appropriate numerical tolerances. Only correct trajectories with valid tool calls, complete termination, and consistent formatting are included in SFT. We also preserve the intermediate entities and values established by the construction program as optional verification targets, without revealing them during solving. The resulting dataset is
\begin{equation}
\mathcal{D}_{\mathrm{ta}}=\bigl\{(x_i,\tau_i,z_i,y_i)\bigr\}_{i=1}^{M_{\mathrm{ta}}},
\label{eq:axion_tool_augmented_data}
\end{equation}
where $z_i=\{v_{i,1},\ldots,v_{i,m_i}\}$ contains the intermediate verification targets associated with the question. Together, the two datasets support a progression from independent scientific reasoning to reasoning grounded in retrieved information and tool use.

\section{Training}
\label{sec:model_training}

\subsection{Overview}
\label{sec:training_overview}

Axion 1.0 follows a three-stage training recipe: two consecutive stages of supervised fine-tuning (SFT), followed by reinforcement learning (RL). The first SFT stage develops self-contained reasoning, the second introduces tool-dependent problem solving, and RL further improves decision-making over the model's own tool-interaction trajectories. We use the data constructed in Section~\ref{sec:data_construction} without additional mid-training, direct preference optimization (DPO), or on-policy distillation (OPD). The infrastructure supporting long-trajectory SFT and asynchronous RL is described in Section~\ref{sec:infra}.

\subsection{Two-Stage Supervised Fine-Tuning}
\label{sec:two_stage_sft}

We train the model to reason before requiring it to retrieve and integrate external information. In \textbf{Stage I}, we fine-tune on the self-contained reasoning dataset $\mathcal{D}_{\mathrm{sc}}$ (Section~\ref{sec:stage1}), where solutions may use Python but not search or webpage visits. In \textbf{Stage II}, we continue training from the Stage I checkpoint on the tool-dependent dataset $\mathcal{D}_{\mathrm{ta}}$ (Section~\ref{sec:stage2}). With construction-time databases withheld, these trajectories demonstrate how to search for missing facts, inspect retrieved pages, and combine external evidence with reasoning and computation. We analyze the contribution of each training stage in Section~\ref{sec:training_stage_analysis}.

\myparatight{Assistant-only supervision}
For both datasets, each training example contains a question $x$, a trajectory $\tau$, a reference answer $y$, and verification targets $z$ (empty for Stage I). We serialize the complete interaction, including tool observations, but supervise only assistant-generated tokens: reasoning, tool calls, and final answers. System and user messages and tool observations provide context without contributing to the loss. The reference answer and verification targets are used for checking trajectories and are not included in the model input. The SFT objective is
\begin{equation}
\begin{aligned}
\mathcal{L}_{\mathrm{SFT}}(\theta;\mathcal{D}_k)
&=-\sum_{(x,\tau,z,y)\in\mathcal{D}_k} \\
&\quad\sum_{j\in\mathcal{I}_{\mathrm{asst}}(\tau)}
\log\pi_\theta(\tau_j\mid x,\tau_{<j}),
\end{aligned}
\label{eq:axion_sft}
\end{equation}
for $k\in\{\mathrm{sc},\mathrm{ta}\}$, where $\pi_\theta$ denotes the model's conditional token distribution and $\mathcal{I}_{\mathrm{asst}}(\tau)$ indexes assistant-token positions in the interaction history.

\subsection{Reinforcement Learning}
\label{sec:stage3}

SFT teaches the model from successful demonstrations, but does not directly optimize its decisions when generating and acting on its own. Starting from the Stage II checkpoint, we therefore apply group-relative policy optimization (GRPO)~\citep{deepseekai2025deepseekr1} with the \texttt{verl} framework~\citep{sheng2024hybridflow}. The model generates trajectories while interacting with search, webpage visits, scholarly retrieval, and Python tools, with context management supporting long-horizon rollouts. RL tasks come from the programmatic construction pipeline in Section~\ref{sec:stage2}, which provides both verifiable final answers and intermediate targets. This allows us to reward correct outcomes while giving partial credit for meaningful progress on unfinished solutions.

\myparatight{Outcome and process rewards}
For a task with question $x$, reference answer $y$, and verification targets $z$, let $\tau$ be a sampled trajectory. We define $F(\tau;y)\in\{0,1\}$ to indicate whether the final answer is correct, $A(\tau)\in\{0,1\}$ to indicate whether a valid final answer is submitted, and $P(\tau;z)\in[0,1]$ to measure verified intermediate progress. Writing $F$, $A$, and $P$ for these trajectory-specific values, we define the reward as
\begin{equation}
R(\tau;y,z)=
\begin{cases}
1, & F=1,\\
\max\!\left(0,\,0.3P-0.1(1-A)\right), & F=0.
\end{cases}
\label{eq:axion_reward}
\end{equation}
A correct final answer receives the full reward. Otherwise, validated intermediate progress can receive partial credit, while the non-submission penalty discourages trajectories that make progress without producing a final answer.

\myparatight{Verification of intermediate progress}
For an ordered chain of $m\geq 1$ dependent targets $v_1,\ldots,v_m$ recorded in $z$, let $b_j(\tau;z)\in\{0,1\}$ indicate whether the trajectory correctly establishes target $v_j$. We aggregate these checks in dependency order:
\begin{equation}
P(\tau;z)=\frac{1}{m}\sum_{j=1}^{m}\prod_{\ell=1}^{j}b_\ell(\tau;z),
\label{eq:axion_process}
\end{equation}
so a later step earns credit only when all its prerequisites have been verified. For example, if only the first and third checks pass in a three-step chain, then $P(\tau;z)=1/3$. Tasks without intermediate targets receive $P(\tau;z)=0$.

We require intermediate results to be identified or used in the agent's reasoning, rather than merely appearing among retrieved candidates. Targets that require external grounding, such as entity identifiers, must also be supported by tool observations. Verification combines deterministic checks with language-model judgments of semantic equivalence. Numerical tolerances are computed programmatically and passed explicitly to the judge, while rule-based matching extracts relevant trajectory passages for evaluating intermediate steps.

\myparatight{RL task selection}
We select RL questions from the same programmatically constructed task pool, retaining the question $x$, reference answer $y$, and intermediate targets $z$. We exclude questions that overlap with held-out evaluations or have known answer or formatting defects. For each remaining question, we sample four trajectories from the current model and score them using Equation~\ref{eq:axion_reward}. We retain the question only if at least one trajectory reaches the correct final answer and the four rewards have nonzero variance. This favors tasks with a demonstrated solution path and a useful group-relative learning signal.

\section{Evaluation}
\label{sec:evaluation}

We evaluate Axion 1.0 along two dimensions: its performance relative to existing open-weight and proprietary models, and the capability gains associated with our three-stage training recipe (Section~\ref{sec:model_training}). We first present the evaluation protocol and overall results, then analyze successive training checkpoints, base-model choice, and performance across model sizes.

\subsection{Evaluation Setup}
\label{sec:evaluation_setup}

\myparatight{Models}
We evaluate two main Axion 1.0 configurations: 35B-A3B and 122B-A10B, initialized from Qwen3.6-35B-A3B and Qwen3.5-122B-A10B, respectively. We also evaluate their base models and intermediate training checkpoints to examine the effects of successive stages. To assess transfer to smaller models, we train 4B and 9B variants with the two SFT stages, without RL. Our models are evaluated in a ReAct environment supporting web search, webpage visits, scholarly retrieval, and Python execution, subject to benchmark-specific restrictions.

\myparatight{Training settings}
We use full-parameter SFT with sample packing. RL uses asynchronous GRPO with eight rollouts per question and session-level training~\citep{xie2026quest}, where context-condensation events divide each rollout into sessions sharing its reward and group-relative advantage. The training infrastructure and systems-level optimizations behind these runs are described in Section~\ref{sec:infra}, and detailed hyperparameters are provided in Tables~\ref{tab:sft_hyperparameters} and~\ref{tab:rl_hyperparameters} of Appendix~\ref{app:training_hyperparameters}.

\myparatight{Benchmarks}
We report results on six evaluation settings: Humanity's Last Exam (HLE)~\citep{phan2025lastexam}, BioMysteryBench (BMB)~\citep{anthropic2026biomysterybench}, SEAL-0~\citep{pham2025sealqa}, HiPhO~\citep{yu2025hipho}, and the Olympiad and Research splits of FrontierScience~\citep{openai2025frontierscience}. HLE measures expert-level problem solving across disciplines; SEAL-0 emphasizes information seeking; and HiPhO and FrontierScience test advanced scientific reasoning. For BMB, we use the human-difficult subset, which requires analysis of real biological data. Training-stage and model-size comparisons use a separate 130-question text-only HLE subset (\emph{HLE-130}); these scores should not be compared directly with the main HLE results in Table~\ref{tab:main_results}.

\myparatight{Evaluation protocol}
For models evaluated in our harness, including GPT-5.6 Sol~\citep{openai2026gpt56} and Gemini 3.1 Pro~\citep{google2026gemini31pro}, we conduct three runs per benchmark, except BMB, for which we conduct five. We report the best score from a single run, rather than an aggregate over runs, and use each benchmark's grading procedure. Reported score improvements are absolute differences between percentage scores, not relative changes. BMB evaluations allow access only to benchmark-approved websites. Baseline results are drawn from official reports, prior studies, or our evaluations, as distinguished in Table~\ref{tab:main_results}. Because externally reported baselines may differ in tool access or execution settings, cross-source comparisons are indicative rather than strictly controlled.

\subsection{Overall Agentic Performance}
\label{sec:overall_performance}

Table~\ref{tab:main_results} compares Axion 1.0 with open-weight and proprietary models across the six benchmarks. On full HLE, Axion 1.0-35B and Axion 1.0-122B achieve 52.2\% and 59.0\%, respectively, with access to websites such as Hugging Face blocked. Without these access restrictions, the corresponding scores are 59.1\% and 64.8\%.

Compared with the similarly sized Agents-A1, Axion 1.0-35B improves by 4.6\% on HLE (52.2\% vs.\ 47.6\%), 5.8\% on SEAL-0 (62.2\% vs.\ 56.4\%), and 7.0\% on FrontierScience Olympiad (86.0\% vs.\ 79.0\%). On FrontierScience Research, the 35B and 122B models score 56.7\% and 66.7\%, respectively, compared with 40.0\% for Agents-A1 and 26.7\% for GPT-5.5; the 122B model also exceeds GPT-5.6 Sol's 63.3\% by 3.4\%. The corresponding average rubric scores for our models are 6.67 and 7.09. On BMB, Axion 1.0-122B reaches 41.2\%, compared with 47.1\% for GPT-5.6 Sol and 46.1\% for Claude Fable 5.

Increasing the model size from 35B to 122B improves five of the six benchmark scores. The largest gains are on FrontierScience Research (+10.0\%), HLE (+6.8\%), and BMB (+5.9\%), with a smaller gain on HiPhO (+0.5\%). SEAL-0 is the exception, with 62.2\% for the 35B model and 58.6\% for the 122B model. Overall, the results show strong performance across both scientific reasoning and information-seeking tasks, while also highlighting variation across benchmarks.

\begin{table}[t]
\centering
\small
\setlength{\tabcolsep}{4pt}
\renewcommand{\arraystretch}{1.15}

\resizebox{\linewidth}{!}{\begin{tabular}{l c c c c c c}
\toprule
\textbf{Model}
 & \textbf{\begin{tabular}[c]{@{}c@{}}Humanity's\\Last Exam\end{tabular}}
 & \textbf{\begin{tabular}[c]{@{}c@{}}BioMysteryBench\\(human-difficult)\end{tabular}}
 & \textbf{SEAL-0}
 & \textbf{HiPhO}
 & \textbf{\begin{tabular}[c]{@{}c@{}}FrontierScience\\Olympiad\end{tabular}}
 & \textbf{\begin{tabular}[c]{@{}c@{}}FrontierScience\\Research\end{tabular}} \\
\midrule

\multicolumn{7}{l}{\textit{Open-weight models}} \\
Agents-A1~\citep{agentsa1team2026agentsa1}
 & 47.6 & \TBD & 56.4 & 46.4 & 79.0 & 40.0 \\
Kimi-K2.6~\citep{moonshot2026kimik26}
 & 54.0 & \TBD & 50.5$^\dagger$ & 41.1$^\dagger$ & 73.0$^\dagger$ & 17.9$^\dagger$ \\
DeepSeek-V4-Pro (Preview)~\citep{deepseek2026v4}
 & 48.2 & \TBD & 55.0$^\dagger$ & 38.7$^\dagger$ & 76.0$^\dagger$ & 13.3$^\dagger$ \\
Apodex 1.1 (ReAct)~\citep{apodex2026apodex11}
 & 53.2 & 23.5 & \TBD & \TBD & \TBD & 55.0 \\
Apodex 1.1 (Agent Team)~\citep{apodex2026apodex11}
 & 56.1 & 35.3 & \TBD & \TBD & \TBD & 63.3 \\

\midrule

\multicolumn{7}{l}{\textit{Proprietary models}} \\
GPT-5.5~\citep{openai2026gpt55} & 52.2 & \TBD & 42.3$^\dagger$ & 43.3$^\dagger$ & 78.0$^\dagger$ & 26.7$^\dagger$ \\
GPT-5.6 Sol~\citep{openai2026gpt56}
 & \TBD & 47.1$^{*}$ & 53.2$^{*}$ & 42.9$^{*}$ & 79.0$^{*}$ & 63.3$^{*}$ \\
GPT-6 Astra~\citep{openai2026gpt6astra}
 & 57.2 & \TBD & \TBD & \TBD & \TBD & \TBD \\
Muse-Spark-1.0~\citep{meta2026musespark}
 & 58.0 & \TBD & \TBD & \TBD & \TBD & 38.0 \\
Gemini 3.1 Pro~\citep{google2026gemini31pro}
 & 51.4 & 35.3$^{*}$ & 53.2$^{*}$ & 38.5$^{*}$ & 85.0$^{*}$ & 26.7$^{*}$ \\
Claude Opus 5~\citep{anthropic2026claudeopus5}
 & 63.6 & 49.4 & \TBD & \TBD & \TBD & \TBD \\
Claude Fable 5~\citep{anthropic2026claudefable5}
 & 63.8 & 46.1 & \TBD & \TBD & \TBD & \TBD \\

\midrule

\rowcolor{tabrow}
\textbf{Axion 1.0-35B}
 & 52.2 & 35.3 & 62.2 & 45.7 & 86.0 & 56.7 \\

\rowcolor{tabrow}
\textbf{Axion 1.0-122B}
 & 59.0 & 41.2 & 58.6 & 46.2 & 87.0 & 66.7 \\

\bottomrule
\end{tabular}}

\caption{\textbf{Overall performance of Axion 1.0.}
Benchmark scores are reported as percentages. Shaded rows indicate our models, and \TBD{} indicates an unreported result. Axion 1.0 scores are evaluated by us. For baselines, $^\dagger$ indicates evaluation by Agents-A1 or Apodex rather than the model provider, while $^{*}$ indicates evaluation by us; unmarked scores are from official reports. Cross-source comparisons may involve different execution settings. Axion's HLE scores use restricted website access; results without these restrictions are reported in the text. FrontierScience Research rubric scores are reported separately in the text.}
\label{tab:main_results}
\end{table}

\subsection{Contribution of Each Training Stage}
\label{sec:analysis}
\label{sec:training_stage_analysis}

Table~\ref{tab:stage_ablation} follows each main model from its base checkpoint through SFT Stage I, SFT Stage II, and RL. These sequential checkpoint comparisons show how performance changes as training progresses; they are not component-removal ablations.

\myparatight{Stage I: self-contained reasoning}
The first SFT stage yields its largest improvements on tasks requiring multi-step scientific reasoning. For the 35B model, HLE-130 increases from 47.7\% to 59.2\%, HiPhO from 38.6\% to 45.9\%, and FrontierScience Olympiad from 71.0\% to 83.0\%. FrontierScience Research also increases from 26.7\% to 48.3\%. The 122B model shows similar gains: HLE-130 increases from 40.8\% to 68.5\% and FrontierScience Olympiad from 65.0\% to 85.0\%. These improvements are consistent with Stage I strengthening the reasoning and computation emphasized during data construction. They are not universal, however: the 35B model's BMB score decreases from 23.5\% to 17.6\% at this stage.

\myparatight{Stage II: tool-dependent reasoning}
Adding tool-use SFT produces particularly clear gains on tasks involving external information and data analysis. For the 35B model, BMB increases from 17.6\% to 35.3\%, FrontierScience Research from 48.3\% to 58.3\%, SEAL-0 from 50.5\% to 53.2\%, and HLE-130 from 59.2\% to 66.2\%. For the 122B model, BMB rises from 29.4\% to 35.3\% and FrontierScience Research from 56.7\% to 65.0\%, while HLE-130 remains at 68.5\%. These results support the intended progression from self-contained reasoning to finding, integrating, and computing with external information.

\myparatight{Reinforcement learning}
RL further improves HLE-130 and SEAL-0 for both model sizes. For the 35B model, scores increase from 66.2\% to 70.0\% on HLE-130 and from 53.2\% to 62.2\% on SEAL-0; for the 122B model, they increase from 68.5\% to 74.6\% and from 53.2\% to 58.6\%, respectively. RL also improves BMB for the 122B model (35.3\% to 41.2\%). Its effects on scientific reasoning benchmarks vary by model size: FrontierScience Research declines from 58.3\% to 56.7\% for 35B but improves from 65.0\% to 66.7\% for 122B. HiPhO decreases slightly for 35B (46.8\% to 45.7\%) but improves slightly for 122B (45.8\% to 46.2\%). Thus, RL provides additional gains on agentic tasks, while its benefits on scientific reasoning benchmarks are more consistent for the larger model.

Overall, the checkpoint results indicate that much of the capability gain arises during the two SFT stages, with RL providing further improvements on selected benchmarks. This performance is achieved without dedicated mid-training, DPO, or OPD, consistent with the simple training recipe described in Section~\ref{sec:model_training}. Appendix~\ref{app:case_studies} complements these results with selected trajectory comparisons illustrating self-checking after Stage I, evidence-grounded retrieval after Stage II, and the synthesis of conflicting information after RL.

\begin{table}[t]
\centering
\small
\setlength{\tabcolsep}{3.5pt}
\renewcommand{\arraystretch}{1.15}

\resizebox{\linewidth}{!}{\begin{tabular}{l c c c c c c}
\toprule
\textbf{Training stage}
 & \textbf{HLE-130}
 & \textbf{\begin{tabular}[c]{@{}c@{}}BioMysteryBench\\(human-difficult)\end{tabular}}
 & \textbf{SEAL-0}
 & \textbf{HiPhO}
 & \textbf{\begin{tabular}[c]{@{}c@{}}FrontierScience\\Olympiad\end{tabular}}
 & \textbf{\begin{tabular}[c]{@{}c@{}}FrontierScience\\Research\end{tabular}} \\
\midrule

\multicolumn{7}{l}{\textbf{Axion 1.0-35B-A3B}} \\

Qwen3.6-35B-A3B & 47.7 & 23.5 & 44.1 & 38.6 & 71.0 & 26.7 \\

\,+ SFT Stage I & 59.2 & 17.6 & 50.5 & 45.9 & 83.0 & 48.3 \\

\,+ SFT Stage II & 66.2 & 35.3 & 53.2 & 46.8 & 84.0 & 58.3 \\

\rowcolor{tabrow}
\,+ RL (Axion 1.0) & 70.0 & 35.3 & 62.2 & 45.7 & 86.0 & 56.7 \\

\midrule

\multicolumn{7}{l}{\textbf{Axion 1.0-122B-A10B}} \\

Qwen3.5-122B-A10B & 40.8 & 17.6 & 47.7 & 30.6 & 65.0 & 40.0 \\

\,+ SFT Stage I & 68.5 & 29.4 & 52.3 & 45.6 & 85.0 & 56.7 \\

\,+ SFT Stage II & 68.5 & 35.3 & 53.2 & 45.8 & 86.0 & 65.0 \\

\rowcolor{tabrow}
\,+ RL (Axion 1.0) & 74.6 & 41.2 & 58.6 & 46.2 & 87.0 & 66.7 \\

\bottomrule
\end{tabular}}

\caption{\textbf{Performance across successive training stages.} Each block tracks one base model through SFT Stage I, SFT Stage II, and RL; shaded rows correspond to the final Axion 1.0 models. HLE uses the 130-question text-only subset. Scores are reported as percentages and follow the protocol in Section~\ref{sec:evaluation_setup}.}
\label{tab:stage_ablation}
\end{table}

\subsection{Additional Analysis}
\label{sec:other_ablations}

\begin{figure}[t]
\centering
\includegraphics[width=\linewidth]{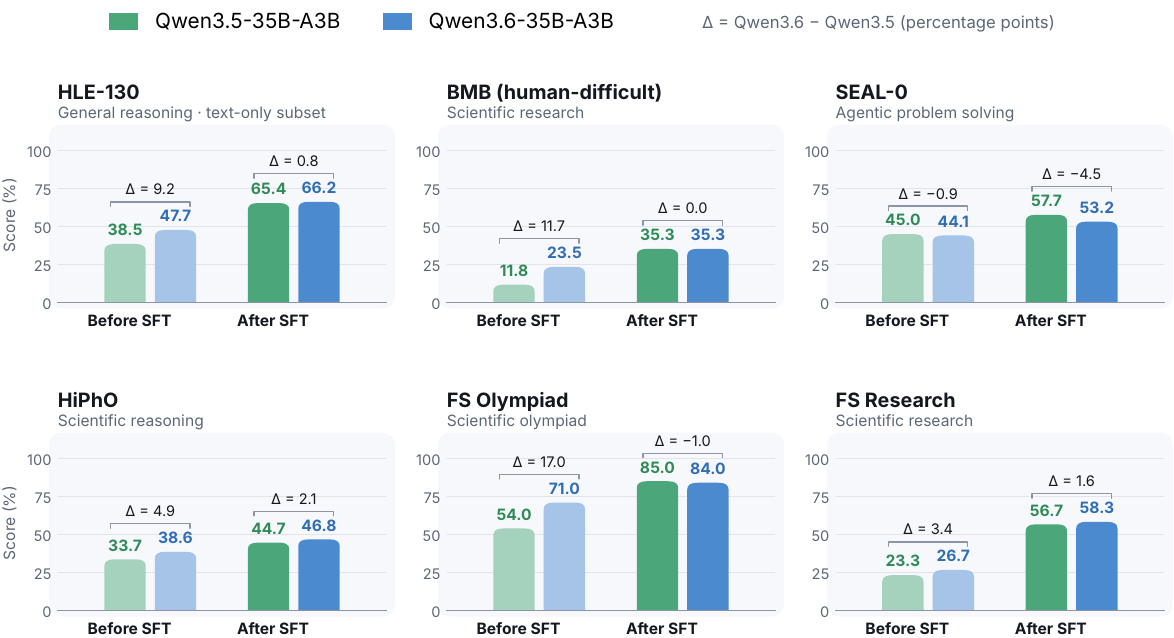}
\caption{\textbf{Effect of the base model.} Two 35B-A3B initializations are evaluated before and after the same two-stage SFT recipe. HLE results use the 130-question text-only subset. All scores follow the protocol in Section~\ref{sec:evaluation_setup}.}
\label{fig:base_model}
\end{figure}

\myparatight{Effect of the base model}
\label{sec:base_model_effect}
Figure~\ref{fig:base_model} compares Qwen3.5-35B-A3B and Qwen3.6-35B-A3B before and after the same two SFT stages. Qwen3.5 initially trails Qwen3.6 on five of the six benchmarks, but the gap narrows substantially after SFT. On HLE-130, the difference decreases from 9.2\% to 0.8\%; on BMB, both models reach 35.3\%. The relationship is not identical on every benchmark: after SFT, Qwen3.5 obtains a higher SEAL-0 score (57.7\% vs.\ 53.2\%). These results suggest that the two-stage SFT recipe is effective for both initializations, rather than relying exclusively on the stronger starting checkpoint.

\myparatight{Effect of SFT across model scale}
\label{sec:model_size_scaling}
Table~\ref{tab:scaling} reports results for 4B, 9B, 35B, and 122B models. Among the smaller models, which receive the same two SFT stages without RL, the 9B model outperforms the 4B model on all six benchmarks. In particular, HLE-130 improves from 53.1\% to 63.1\%, BMB from 11.8\% to 23.5\%, SEAL-0 from 56.8\% to 61.3\%, and FrontierScience Research from 45.0\% to 55.0\%. This indicates that the constructed SFT data are effective at both smaller model sizes. The 35B and 122B results provide a broader view of final-model performance, but should not be interpreted as a controlled size-scaling experiment because these models additionally undergo RL and use different base architectures.

\myparatight{Effect of RL across model scales} Table~\ref{tab:stage_ablation} compares the 35B and 122B models before and after RL. For \textbf{Axion 1.0-35B}, RL yields its largest gains on SEAL-0 (+9.0\%) and HLE-130 (+3.8\%), with a smaller improvement on FrontierScience Olympiad (+2.0\%). BMB remains unchanged, while performance decreases slightly on full HiPhO ($-$1.1\%) and FrontierScience Research ($-$1.6\%). For \textbf{Axion 1.0-122B}, RL yields its largest gains on HLE-130 (+6.1\%), BMB (+5.9\%), and SEAL-0 (+5.4\%), with smaller improvements on full HiPhO and both FrontierScience subsets. Across these two models, RL consistently improves HLE-130, SEAL-0, and FrontierScience Olympiad, while its effects on the remaining benchmarks vary. These comparisons describe the gains for each model under its training configuration, rather than isolating model scale as their cause.

\myparatight{Effect of rollout routing replay}
On Qwen3.5-122B-A10B, rollout routing replay (R3)~\citep{ma2025stabilizingmoereinforcementlearning} improves rollout--trainer consistency: the mean sequence-average log-probability gap decreases from 0.044 to 0.042, and the maximum sequence-average gap decreases from 1.799 to 1.282; mean gradient norm decreases from 1.998 to 1.922, mean PPO KL from 0.0413 to 0.0396, and mean clipping fraction from 3.86\% to 3.68\%. These improvements accompany stronger performance on HLE-130 (74.6\% vs.\ 72.3\%), FrontierScience Olympiad (87.0\% vs.\ 85.0\%), and FrontierScience Research (66.7\% vs.\ 61.7\%). HiPhO performance remains similar (46.2\% vs.\ 46.3\%), whereas SEAL-0 decreases from 63.1\% to 58.6\%. Overall, R3 improves rollout--trainer consistency and yields gains on several reasoning-heavy benchmarks, although the downstream benefits vary by benchmark. These results provide encouraging evidence for routing replay. However, the log-probability diagnostics also reflect asynchronous policy staleness and the off-policy effects of context condensation, limiting causal attribution of the benchmark differences to R3 alone.

\begin{table}[t]
\centering
\small
\setlength{\tabcolsep}{3.5pt}
\renewcommand{\arraystretch}{1.15}

\resizebox{\linewidth}{!}{\begin{tabular}{l c c c c c c}
\toprule
\textbf{Model}
 & \textbf{HLE-130}
 & \textbf{\begin{tabular}[c]{@{}c@{}}BioMysteryBench\\(human-difficult)\end{tabular}}
 & \textbf{SEAL-0}
 & \textbf{HiPhO}
 & \textbf{\begin{tabular}[c]{@{}c@{}}FrontierScience\\Olympiad\end{tabular}}
 & \textbf{\begin{tabular}[c]{@{}c@{}}FrontierScience\\Research\end{tabular}} \\
\midrule

Axion 1.0-4B & 53.1 & 11.8 & 56.8 & 42.3 & 81.0 & 45.0 \\

Axion 1.0-9B & 63.1 & 23.5 & 61.3 & 43.5 & 83.0 & 55.0 \\

Axion 1.0-35B & 70.0 & 35.3 & 62.2 & 45.7 & 86.0 & 56.7 \\

Axion 1.0-122B & 74.6 & 41.2 & 58.6 & 46.2 & 87.0 & 66.7 \\

\bottomrule
\end{tabular}}

\caption{\textbf{Performance across model sizes.} The 4B and 9B models are trained with two SFT stages, while the 35B-A3B and 122B-A10B models additionally undergo RL. HLE uses the 130-question text-only subset. Scores are reported as percentages and follow the protocol in Section~\ref{sec:evaluation_setup}.}
\label{tab:scaling}
\end{table}

\section{Training Infrastructure}
\label{sec:infra}

We develop separate infrastructure for full-parameter SFT on long agent trajectories and asynchronous RL with tool-interacting rollouts. Both are designed to train large mixture-of-experts (MoE) models efficiently while preserving the interactions needed for agentic learning.

\subsection{Supervised Fine-Tuning Framework}

Our SFT framework integrates NVIDIA Megatron-FSDP~\citep{zhao2023pytorchfsdpexperiencesscaling} into the Megatron backend of ms-swift~\citep{zhao2024swiftascalablelightweightinfrastructure}. It combines ZeRO-3-style sharding of parameters, gradients, and optimizer states~\citep{rajbhandari2020zeromemoryoptimizationstraining} with Megatron's optimized kernels and parallelism~\citep{shoeybi2020megatronlmtrainingmultibillionparameter}. Unlike session-based training with context condensation, SFT uses each complete agent trajectory as one example, preserving all conversation turns and tool observations without replacing earlier context with summaries. This retains the original interaction history but requires training on sequences approaching 128K tokens.

\myparatight{Parallelism for large MoE models}
We use fully sharded data parallelism (FSDP) rather than pipeline parallelism~\citep{narayanan2021efficientlargescalelanguagemodel}, avoiding pipeline bubbles and stage-to-stage communication that otherwise require carefully tuned schedules~\citep{qi2023zerobubblepipelineparallelism}. FSDP distributes model states across GPUs and overlaps communication with computation. Within MoE layers, we use expert parallelism~\citep{liu2026moeparallelfoldingheterogeneous} instead of tensor parallelism~\citep{shoeybi2020megatronlmtrainingmultibillionparameter}: keeping expert computations intact avoids splitting small expert matrix multiplications and introducing additional all-reduces. We set the tensor-parallel degree to 1 and the expert-parallel degree to 8, keeping expert dispatch and combination within a single NVLink domain.

\myparatight{Context parallelism for long trajectories}
FSDP alone does not shard all activations. For Qwen3.5-122B-A10B, the remaining activation memory exceeds 1.4 MiB per token, including MoE dispatch buffers, gated-delta-net (GDN) states, and vocabulary logits. We therefore partition each sequence across GPUs, using Ring Attention~\citep{liu2023ringattentionblockwisetransformers} for full-attention layers and a Ulysses-style all-to-all~\citep{jacobs2023deepspeedulyssesoptimizationsenabling} for GDN layers. On 128K-token sequences, context-parallel degree 4 and micro-batch size 2 provide the highest throughput in our tests.

\myparatight{Training cost}
All SFT runs use a 128K context window on 32 GPUs. For Qwen3.5-122B-A10B, Stage I processes 841M tokens in 3\,h 32\,min and Stage II takes 3\,h 58\,min, totaling 7\,h 30\,min (240 GPU-hours), with peak memory of 113 GiB per GPU. For Qwen3.6-35B-A3B, Stage I processes 657M tokens in 1\,h 2\,min and Stage II takes 1\,h 46\,min, totaling 2\,h 48\,min (approximately 90 GPU-hours), with peak memory of 101 GiB per GPU.

\subsection{Reinforcement Learning Framework}

Our RL infrastructure builds on verl's fully asynchronous policy module and separates rollout generation, policy optimization, and auxiliary services onto distinct GPU pools~\citep{sheng2024hybridflow}. Rollout workers use vLLM to execute the agent loop with search, webpage visits, Python, and QUEST-based context condensation. Trainer workers run the Megatron-Core actor and reference model. A separate service node hosts two Qwen3.8-27B judges for open-ended and citation rewards and six Qwen3.5-9B instances for webpage summarization and context condensation. For Qwen3.5-122B-A10B, we use ten tensor-parallel-4 rollout engines on 40 GPUs and a 16-GPU trainer (tensor parallelism 2, expert parallelism 8, with optimizer states offloaded to CPU), alongside the service node.

\myparatight{Asynchronous rollout and optimization}
Rollout and training workers run concurrently on separate GPUs. Rollout engines continuously generate trajectories and enqueue completed results; the trainer starts an update once it has a batch of 24 prompts with non-degenerate rewards. Every two optimizer steps, updated weights are transferred to the rollout engines through NCCL, taking 45--60 seconds for the 122B model. In-flight rollouts continue with the updated weights, while trajectories more than 11 parameter versions old are discarded. This separation avoids the inference-memory contention of colocating rollout and training workers and improves throughput for the 122B model. Following QUEST, context-condensation boundaries divide RL rollouts into sessions. Sessions from the same rollout share its reward and group-normalized advantage, and the training loss is normalized per session.

\myparatight{Token consistency with Token-In-Token-Out (TITO)}
Reconstructing sessions by rendering messages and re-tokenizing them can change token IDs because byte-pair encoding (BPE) is sensitive to token boundaries. This creates a mismatch between the sequence used for rollout and the one used for training. We avoid this with TITO~\citep{sheng2024hybridflow}: the agent loop exchanges token IDs directly, appends generated and tool-output tokens to a shared token buffer, and supplies the trainer with that buffer and a mask of policy-generated positions. After context condensation, the new prefix is tokenized only once at session start and joined to subsequent tokens without a decode--re-encode cycle.

\myparatight{Expert-routing consistency with Rollout Routing Replay (R3)}
For MoE models, vLLM and Megatron can select different experts for the same tokens even with identical weights. R3 addresses this mismatch by recording the selected experts for each token and MoE layer during rollout and replaying those decisions in the trainer's forward pass~\citep{ma2025stabilizingmoereinforcementlearning}. The routing records use \texttt{uint8} tensors (approximately 20 MB per sample). On Qwen3.5-122B-A10B, R3 reduces mean gradient norm from 1.998 to 1.922, mean PPO KL from 0.0413 to 0.0396, and mean clipping fraction from 3.86\% to 3.68\%. It also halves the number of intervals with gradient norm above 5. However, the largest gradient spike increases from 21.1 to 26.8 and reward standard deviation rises by 15\%, indicating improved average consistency rather than elimination of occasional instability.

\section{Related Work}
\label{sec:related_work}

\myparatight{Agent foundation models}
Recent models increasingly integrate reasoning, coding, and tool use as general-purpose capabilities, as illustrated by Kimi K2~\citep{kimiteam2025kimik2} and GLM-4.5~\citep{zeng2025glm45}. Research-oriented agents extend this direction to sustained information gathering and interaction with external environments. MiroThinker-1.7~\citep{miromind2026mirothinker17} combines agentic mid-training, SFT, preference optimization, and RL; Agents-A1~\citep{agentsa1team2026agentsa1} uses long-horizon supervision and domain-routed on-policy distillation; and Apodex 1.1~\citep{apodex2026apodex11} investigates executable environments and agent coordination. These systems explore different approaches to building capable agents. In contrast, Axion 1.0 studies how far a simpler sequence of two SFT stages and RL can go, without dedicated mid-training, direct preference optimization, or on-policy distillation.

\myparatight{Construction of challenging and verifiable tasks}
Synthetic data has long been used to expand training supervision. Self-Instruct~\citep{wang2023selfinstruct} generates instruction data from model outputs, while Evol-Instruct~\citep{xu2024wizardlm} increases task complexity through iterative transformations. More recent work adapts task difficulty to the learner: Absolute Zero~\citep{zhao2025absolutezero} uses execution-grounded self-play, and Agent0~\citep{xia2025agent0} develops curricula through interaction between a task generator and a tool-using agent. Axion's first data stage similarly uses model feedback, but specifically strengthens questions that the base model solves correctly by targeting its successful solution strategy. This produces harder, self-contained reasoning tasks with independently verified answers.

Tool-use training additionally requires tasks that demand external information while remaining verifiable. TaskCraft~\citep{shi2025taskcraft} expands verifiable multi-tool tasks along depth and breadth, WebShaper~\citep{tao2025webshaper} structures information-seeking complexity, and QUEST~\citep{xie2026quest} constructs research tasks with verifiable rubrics. Axion's second data stage takes a complementary approach: deterministic filtering, unique-entity selection, and computation over structured data produce reference answers before an LLM formulates the questions. With construction-time data withheld and closed-book-solvable questions filtered out, the resulting tasks emphasize retrieval, webpage inspection, and grounded computation.

\myparatight{Reinforcement learning for reasoning and tool use}
DeepSeek-R1~\citep{deepseekai2025deepseekr1} demonstrates the potential of RL to improve reasoning. Search-R1~\citep{jin2025searchr1} extends RL to reasoning interleaved with retrieval, DeepResearcher~\citep{zheng2025deepresearcher} trains agents through open-web interaction, and AgentGym-RL~\citep{xi2025agentgymrl} studies multi-turn learning across environments. Together, these works show how RL can improve sequential decisions, not only final responses. Axion applies RL after its two complementary SFT stages, building on supervised reasoning and tool-use trajectories to improve long-horizon interaction.

\myparatight{Training--inference consistency in RL}
Agentic RL typically generates trajectories with an inference engine and computes policy updates in a separate training backend. Differences in numerical execution, expert routing, or tokenization can cause the two systems to assign different probabilities to the same trajectory, complicating policy optimization. One line of work addresses this mismatch through truncated importance sampling~\citep{yao2025offpolicy}, sequence-aware masking~\citep{li2026trustregionmaskinglonghorizon}, and analyses of the conditions under which token-level corrections remain valid~\citep{zheng2025stabilizingreinforcementlearningllms}. Another seeks to reduce the discrepancy itself through expert-routing replay~\citep{ma2025stabilizingmoereinforcementlearning}, numerical-format choices~\citep{qi2025defeatingtraininginferencemismatchfp16}, and precision-aware optimization~\citep{zhang2026precisiontraininginferencemismatchoptimization}. Controlled studies further examine the instability caused by numerical mismatch~\citep{zhong2026diagnosingtraininginferencemismatch}. In long agent trajectories, reconstructing messages and tool observations can introduce an additional tokenization mismatch. Axion addresses these two sources directly: Token-In-Token-Out (TITO) preserves the rollout token sequence across training, while Rollout Routing Replay (R3) aligns MoE expert choices between the inference and training engines.

\myparatight{Open and reproducible training recipes}
Reproducibility requires more than releasing model weights: data sources, construction procedures, and training configurations also determine the resulting capabilities. OpenThoughts~\citep{guha2025openthoughts} develops open reasoning-data recipes, and OpenThoughts-Agent~\citep{raoof2026openthoughtsagent} investigates and releases agentic data-curation procedures. DR Tulu~\citep{shao2025drtulu} provides an open RL framework for research agents, while QUEST~\citep{xie2026quest} releases models, data, and training scripts. Axion follows this direction by making its three-stage recipe---including data generation, filtering, verification, prompts, and training procedures---available for inspection and reproduction. Its emphasis is on obtaining broad agentic capabilities with a compact, reproducible pipeline.

\section{Conclusion}
\label{sec:conclusion}

We present Axion 1.0, demonstrating that strong agentic capabilities can be developed through a simple yet effective, data-centric training paradigm.
Our approach progressively equips models with the ability to reason over complex problems, acquire and utilize external knowledge, and refine their capabilities through reinforcement learning.
Across challenging evaluations, Axion 1.0 demonstrates strong scientific reasoning and tool-use capabilities, highlighting the effectiveness of this streamlined approach.
More broadly, our findings suggest that advancing agent foundation models does not necessarily require increasingly complex training pipelines; carefully constructed data and well-designed learning objectives can be equally powerful.
By releasing our data construction pipelines and training code, we hope to provide a reproducible and accessible foundation for future research toward more capable and general-purpose AI agents.

\clearpage
\bibliographystyle{plainnat}
\bibliography{custom}

\clearpage
\beginappendix
\section{Training Hyperparameters}
\label{app:training_hyperparameters}

Tables~\ref{tab:sft_hyperparameters} and~\ref{tab:rl_hyperparameters} summarize the SFT and RL hyperparameters, respectively.
Both SFT stages use the same optimization settings, with Stage II initialized from the Stage I checkpoint.

\begin{table}[htbp]
\centering
\small
\setlength{\tabcolsep}{7pt}
\renewcommand{\arraystretch}{1.13}
\begin{tabular}{lc}
\toprule
\textbf{Hyperparameter} & \textbf{Value} \\
\midrule
Learning rate & $3\times10^{-6}$ \\
Learning-rate schedule & Constant \\
Warmup steps & 0 \\
Adam $(\beta_1,\beta_2)$ & $(0.9,0.95)$ \\
Weight decay & 0.1 \\
Gradient clipping & 1.0 \\
Training precision & BF16 \\
Epochs per stage & 3 \\
Maximum sequence length & 128K \\
Global batch size & 8 \\
\bottomrule
\end{tabular}
\caption{\textbf{SFT hyperparameters.} Settings apply to both supervised stages. Batch size counts sequences per optimizer update.}
\label{tab:sft_hyperparameters}

\vspace{1em}
\begin{tabular}{lc}
\toprule
\textbf{Hyperparameter} & \textbf{Value} \\
\midrule
Learning rate & $2\times10^{-6}$ \\
Learning-rate schedule & Constant \\
Warmup steps & 5 \\
Adam $(\beta_1,\beta_2)$ & $(0.9,0.999)$ \\
Weight decay & 0.1 \\
Gradient clipping & 1.0 \\
Training precision & BF16 \\
Rollouts per training question & 8 \\
Rollout temperature & 1.0 \\
Maximum session prompt length & 12,000 \\
Maximum session response length & 49,152 \\
Lower / upper clipping range & $0.2\,/\,0.28$ \\
KL-loss coefficient & 0.001 \\
Entropy coefficient & 0 \\
Parameter-sync interval (updates) & 2 \\
Async staleness threshold & 11 \\
\bottomrule
\end{tabular}
\caption{\textbf{RL hyperparameters.} The response-length limit applies to each session rather than the entire multi-session rollout. The eight rollouts used during training are separate from the four offline attempts used for task selection.}
\label{tab:rl_hyperparameters}
\end{table}

\clearpage
\begingroup
\color{black}
\section{Case Studies Across Training Stages}
\label{app:case_studies}

We present three selected evaluation cases illustrating the progression from self-contained reasoning to evidence-grounded problem solving and the synthesis of conflicting information. Each case compares an earlier checkpoint with a later one on the same question. The examples complement the aggregate results in Table~\ref{tab:stage_ablation}: they describe particular solution trajectories rather than establishing that a behavior is absent from every earlier-stage rollout or isolating the causal contribution of a training component.

\begin{table}[htbp]
\centering
\small
\setlength{\tabcolsep}{5pt}
\renewcommand{\arraystretch}{1.2}
\begin{tabular}{@{}p{0.23\linewidth}p{0.27\linewidth}p{0.43\linewidth}@{}}
\toprule
\textbf{Transition} & \textbf{Task} & \textbf{Capability illustrated} \\
\midrule
Base $\rightarrow$ SFT Stage I & Cube-surface geometry & Detecting a flawed numerical approximation and revising the solution method. \\
SFT Stage I $\rightarrow$ Stage II & Philosophical attribution & Retrieving a passage that directly supports the requested attribution. \\
SFT Stage II $\rightarrow$ RL & Ride-booking application & Separating competing usage metrics and resolving them against the question's criteria. \\
\bottomrule
\end{tabular}
\caption{\textbf{Overview of the selected case studies.} The comparisons concern selected 35B trajectories; the discussion below identifies the relevant behavior and answer in each case.}
\label{tab:case_study_overview}
\end{table}

\subsection{Base to Stage I: Reasoning and Self-Checking}
\label{app:case_stage1}

\begin{specbox}
\color{black}
\textbf{Task: geodesic distance on a cube.}
Let $S$ be the surface of a cube of side length $r$, and let $P$ be the midpoint of an edge. The locus $C$ consists of points whose shortest distance to $P$ along $S$ is exactly $r$. Find $|C|/(2\pi r)$ as a whole-number percentage.
\end{specbox}

\myparatight{Earlier behavior}
The case-study record reports that Qwen3.6-35B-A3B attempted to unfold the cube, but mishandled the coordinate transformations and the assignment of circular arcs to faces. It obtained a total length of $2\pi r$ and answered $100\%$, without checking the derivation computationally.

\myparatight{Stage I behavior}
The selected Stage I trajectory first constructs a grid approximation and applies Dijkstra's algorithm to estimate distances along the surface. It then tests this solver against a configuration with a known answer. For $r=1$, take
\[
P=(0,0,1/2),\qquad Q=(0,1/4,1).
\]
Both points lie on one face, so the straight segment on that face has length
\[
 d(P,Q)=\sqrt{(1/4)^2+(1/2)^2}\approx0.559.
\]
The grid solver instead returns approximately $0.604$, and grid refinement does not remove the discrepancy. The trajectory identifies the problem: movement restricted to grid edges produces a longer polyline rather than the straight surface path. It consequently switches to analytical face unfolding, compares alternative paths, and evaluates the resulting arcs. Its final expression is equivalent to
\[
\frac{|C|}{2\pi r}
=1-\frac{\arcsin(3/4)}{\pi}
\approx0.73005,
\]
yielding the correct whole-number answer, $73\%$.

\myparatight{Capability illustrated}
The distinguishing behavior is the use of a known geometric relation to challenge a computed result. Python supplies a candidate calculation, but the trajectory checks its validity, diagnoses a representation error, and changes method before committing to an answer.

\subsection{Stage I to Stage II: Retrieval and Use of Evidence}
\label{app:case_stage2}

\begin{specbox}
\color{black}
\textbf{Task: attributing a philosophical distinction.}
Which philosopher of the Lviv--Warsaw School decomposes the notion of a hallmark into symptoms and signals?
\end{specbox}

\myparatight{Earlier behavior}
In the selected 35B Stage I rollout, the model searches for material on the school's theory of signs and attributes the distinction to its founder, Kazimierz Twardowski. The retrieved material is related to the topic, but the trajectory does not establish that this particular distinction belongs to him.

\myparatight{Stage II behavior}
The selected 35B Stage II rollout searches for the terms of the distinction and opens material from \emph{Studia Semiotyczne}. It locates a passage discussing Izydora D\k{a}mbska's treatment of indications, including their division into symptoms and signals. It connects this passage to the surrounding discussion of D\k{a}mbska's semiotic work and answers \textbf{Izydora D\k{a}mbska}.

\myparatight{Capability illustrated}
Both checkpoints can invoke retrieval tools; the difference in these trajectories lies in how the evidence is used. Stage II connects the requested concept, the specific distinction, and the named philosopher, instead of treating general relevance to the school as sufficient support. This case illustrates evidence-grounded attribution rather than tool invocation alone.

\subsection{Stage II to RL: Reconciling Retrieved Information}
\label{app:case_rl}

\begin{specbox}
\color{black}
\textbf{Task: identifying a ride-booking application.}
Which ride-booking application is most widely used in India and offers a wide variety of vehicle choices and prices? The evaluation prompt asks for information applicable to 2024, when the question was created.
\end{specbox}

\myparatight{Earlier behavior}
The selected 35B Stage II trajectory retrieves information about Ola, Uber, and Rapido. Its final response includes active-usage and ride-volume figures favoring Uber over Ola, but also emphasizes Ola's cumulative installs and historical prominence. It presents Ola and Uber side by side and ultimately favors Ola, rather than resolving which evidence answers the combined criteria.

\myparatight{RL behavior}
The selected 35B RL trajectory distinguishes cumulative downloads, active-user measures, ride counts, and vehicle-category coverage. It notes that the interpretation of ``most widely used'' depends on the metric and that reports can disagree. It then evaluates usage together with the requirement for a broad range of vehicles and price points, explicitly choosing \textbf{Uber} as the answer matching both criteria in the recorded evaluation.

\myparatight{Capability illustrated}
The improvement in this comparison is in the synthesis of retrieved information. The selected RL response separates incompatible measures and ties its conclusion to the question's constraints, while retaining the qualification that different definitions of usage can favor different candidates. These are observations about the recorded trajectories, not a claim about the current ride-hailing market.

\endgroup

\end{document}